\documentclass[10pt,twocolumn,letterpaper]{article}

\usepackage{iccv}
\usepackage{times}
\usepackage{epsfig}
\usepackage{graphicx}
\usepackage{amsmath}
\usepackage{amssymb}

\usepackage{ulem}
\usepackage{xcolor}
\usepackage{array}
\usepackage{multirow}
\usepackage{mathtools}
\usepackage{authblk}
\usepackage{paralist}
\usepackage{booktabs}
\usepackage{enumitem}
\usepackage{overpic}

\ificcvfinal\fi
\DeclarePairedDelimiter{\ceil}{\lceil}{\rceil}
\newcommand{\reffig}[1]{Fig.~\ref{#1}}
\newcommand{\reftab}[1]{Tab.~\ref{#1}}
\newcommand{\refsec}[1]{Sec.~\ref{#1}}

\usepackage[breaklinks=true,bookmarks=false]{hyperref}

\iccvfinalcopy 

\def\iccvPaperID{3681} 

\newcommand*{\Scale}[2][4]{\scalebox{#1}{$#2$}}%
\begin{document}

\title{Audio2Gestures: Generating Diverse Gestures from Speech Audio with Conditional Variational Autoencoders}

\author[1]{Jing Li}
\author[2]{Di Kang}
\author[1]{Wenjie Pei}
\author[2]{Xuefei Zhe}
\author[2]{Ying Zhang}
\author[1]{Zhenyu He \thanks{Corresponding author: zhenyuhe@hit.edu.cn}}
\author[2]{Linchao Bao}
\affil[1]{Harbin Institute of Technology, Shenzhen}
\affil[2]{Tencent AI Lab}

\maketitle

\begin{abstract}
Generating conversational gestures from speech audio is challenging due to the inherent one-to-many mapping between audio and body motions. 
Conventional CNNs/RNNs assume one-to-one mapping, and thus tend to predict the average of all possible target motions, resulting in plain/boring motions during inference. 
In order to overcome this problem, we propose a novel conditional variational autoencoder (VAE) that explicitly models one-to-many audio-to-motion mapping by splitting the cross-modal latent code into shared code and motion-specific code. The shared code mainly models the strong correlation between audio and motion (such as the synchronized audio and motion beats), while the motion-specific code captures diverse motion information independent of the audio. 
However, splitting the latent code into two parts poses training difficulties for the VAE model. 
A mapping network facilitating random sampling along with other techniques including relaxed motion loss, bicycle constraint, and diversity loss are designed to better train the VAE.
Experiments on both 3D and 2D motion datasets verify that our method generates more realistic and diverse motions than state-of-the-art methods, quantitatively and qualitatively.
Finally, we demonstrate that our method can be readily used to generate motion sequences with user-specified motion clips on the timeline.
Code and more results are at  \url{https://jingli513.github.io/audio2gestures}.
\end{abstract}


\vspace{-0.5cm}
\section{Introduction}
\vspace{-0.2cm}
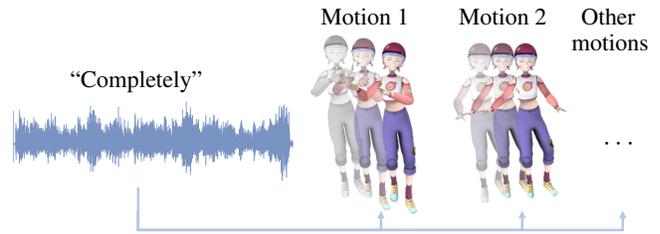
\begin{figure}[!t]
    \centering
    \begin{overpic}[trim=2.7cm 6cm 2cm 5.5cm,clip,width=1\linewidth,grid=false]{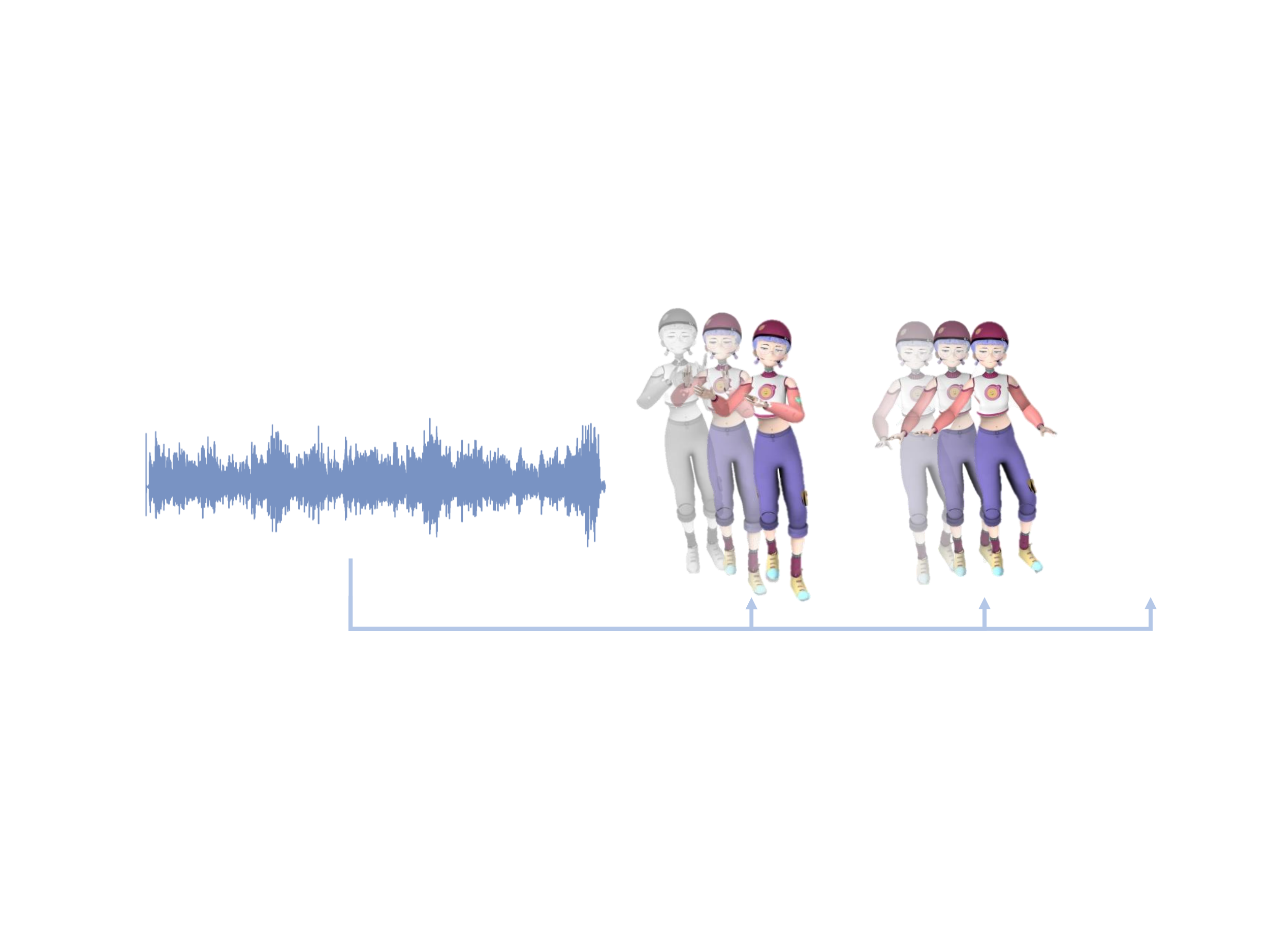}
    \put(10,25){\small ``Completely"}
    \put(50,35){\small Motion 1}
    \put(72,35){\small Motion 2}
    \put(91.5,35){\small Other}
    \put(90,31){\small motions}
    \put(95,15){$\cdots$}
    \end{overpic}\vspace{-6pt}
   \caption{Illustration of the existence of one-to-many mapping between audio and motion in Trinity dataset~\cite{IVA:2018}. Different gestures are performed when the subject says ``completely''.Similar phenomena broadly exist in co-speech gestures. The character used for demonstration is from Mixamo~\cite{mixamo}.}
\label{demo}
\end{figure}
In the real world, co-speech gestures help express oneself better, and in the virtual world, it makes a talking avatar act more vividly. 
Attracted by these merits, there has been a growing demand for generating realistic human motions for given audio clips recently. This problem is very challenging because of the complicated one-to-many relationship between audio and motion.
A speaker may act different gestures when speaking the same words due to different mental and physical states.
\par
Existing algorithms developed for audio to body dynamics have some obvious limitations.
For example, \cite{Ginosar2019} adapts a fully convolutional neural network to co-speech gesture synthesis tasks.
Nevertheless, their model tends to predict averaged motion and thus generates motions lacking diversity.
This is due to the underlying one-to-one mapping assumption of their model, which ignores that the relationship between speech and co-speech gesture is one-to-many in nature.
Under such an overly simplified assumption, the model has no choice but to learn the averaged motion when several motions match almost the same audio clips in order to minimize the error. 
The above evidence inspires us to study whether or not explicitly modeling this multimodality improves the overall motion quality.
To enhance the regression capability, we introduce an extra motion-specific latent code. 
With this varying {\it full} latent code, which contains the same shared code and varying motion-specific code, the decoder can regress different motion targets well for the same audio, achieving one-to-many mapping results.
Under this formulation, the shared code extracted from audio input serves as part of the control signal. 
The motion-specific code further modulates the audio-controlled motion, enabling multimodal motion generation.

Although this formulation is straightforward, it is not trivial to make it work as expected.
Firstly, there exists an easy degenerated solution since the motion decoder could utilize only the motion-specific code to reconstruct the motion.
Secondly, we need to generate the motion-specific code since we do not have access to the target motion during inference.
Our solution to the aforementioned problems is providing {\it random noise} to the motion-specific code so that the decoder has to utilize the deterministic information contained in the shared code to reconstruct the target.
\par
But under this circumstance, it is unsuitable for forcing the motion decoder to reconstruct the exact original target motion anymore.
So a {\it relaxed motion loss} is proposed to apply to the motions generated with random motion-specific code.
Specifically, it only penalizes the joints deviating from their targets larger than a threshold.
This loss encourages the motion-specific code to tune the final motion while respecting the shared code's control.
\par
Our contributions can be summarized as:
\begin{compactitem}
\item We present a co-speech gesture generation model whose latent space is split into shared code and motion-specific code to better regress the training data and generate diverse motions.

\item We utilize random sampling and a relaxed motion loss to avoid degeneration of the proposed network and enable the model to generate multimodal motions.
\item The effectiveness of the proposed method has been verified on 3D and 2D gesture generation tasks by comparing it with several state-of-the-art methods.
\item The proposed method is suitable for motion synthesis from annotations since it can well respect the pre-defined actions in the timeline by simply using their corresponding motion-specific code.
\end{compactitem}
\section{Related Work}
\noindent\textbf{Audio to body dynamics.}
Early methods generate human motion for specified audio input by blending motion clips chosen from a motion database according to hidden Markov model~\cite{Levine2009} or conditional random fields~\cite{levine2010gesture}. Algorithms focusing on selecting motion candidates from a pre-processed database usually cannot generate motions out of the database and does not scale to large databases.
\par
Recently, deep generative models, such as VAEs~\cite{kingma2014auto} and GANs~\cite{goodfellow2014generative}, have achieved great success in generating realistic images, as well as human motions~\cite{yan2018mt,motioninpainting,motionvae}.
For example,~\cite{shlizerman2018audio} utilizes a classic LSTM to predict the body movements of a person playing the piano or violin given the sound of the instruments. 
However, the body movements of a person playing the piano or violin show regular cyclic pattern and are usually constrained within a small pose space.
\par
In contrast, generating co-speech gestures is more challenging in the following two aspects -- the motion to generate is more complicated and the relationship between the speech and motion is more complicated.
As a result, Speech2Gesture~\cite{Ginosar2019} proposes a more powerful fully convolutional network, consisting of a 8-layer CNN audio encoder and a 16-layer 1D U-Net decoder, to translate log-mel audio feature to gestures.
And this network is trained with 14.4 hours of data per individual on average in comparison to 3 hours data in~\cite{shlizerman2018audio}.
Other than greatly enlarged network capacity, this fully convolutional network better avoids the error accumulation problem often faced by RNN-based methods.
However, it still suffers from predicting the averaged motion due to the existence of one-to-many mapping in the training data.
The authors further introduce adversarial loss and notice that the loss helps to improve diversity but degenerates the realism of the outputs.
In contrast, our method avoids learning the averaged motion by explicitly modeling the one-to-many mapping between audio and motion with the help of the extra motion-specific code.
\par
Due to lack of 3D human pose data, the above deep learning based methods~\cite{shlizerman2018audio, Ginosar2019} have only tested 2D human pose data, which are 2D key point locations estimated from videos.
Recently,~\cite{IVA:2018} collects a 3D co-speech gesture dataset named Trinity Speech-Gesture Dataset, containing 244 minutes motion capture (MoCap) data with paired audio, and thus enables deep network-based study on modeling the correlation between audio and 3D motion. 
This dataset has been tested by StyleGestures~\cite{henter2019moglow}, which is a flow-based algorithm~\cite{kingma2018glow, henter2019moglow}. StyleGestures generates 3D gestures by sampling poses from a pose distribution predicted from previous motions and control signals.
However, samples generated by flow-based methods~\cite{kingma2018glow, henter2019moglow} are often not as good as VAEs and GANs.
In contrast, our method learns the mapping between audio and motion with a customized VAE.
Diverse results can be sampled since VAE is a probabilistic generation model.

\vspace{0.1cm}
\noindent\textbf{Human motion prediction.}
There exist many works focus on predicting future motion given previous motion~\cite{motioninpainting,quaternet:2018,yan2018mt}.
It is natural to model sequence data with RNNs~\cite{ERD, jain2016structural,martinez2017human, yan2018mt}.
But~\cite{motioninpainting} has pointed out the RNN-based methods often suffer from error accumulation and thus are not good at predicting long-term human motion. 
So they proposes to use a fully convolutional generative adversarial network and achieves better performance at long-term human motion prediction.
Similarly, we also adopt a fully convolutional neural network since we need to generate long-term human motion.
Specific to 3D human motion prediction, another type of error accumulation happens along the kinematic chain~\cite{quaternet:2018} because any {\it small} joint rotation error propagates to all its descendant joints, e.g. hands and fingers, resulting in {\it considerable} position error especially for the end-effectors (wrists, fingers).
So QuaterNet~\cite{quaternet:2018} optimizes the joint position which is calculated from forward kinematics when predicting long-term motion.
Differently, we optimize the joint rotation and position losses at the same time to help the model learn the joint limitation at the same time.

\vspace{0.1cm}
\noindent\textbf{Multimodal generation tasks.}
Generating data with multimodality has received increasing interests in various tasks, such as image generation~\cite{huang2018multimodal, bicyclegan}, motion generation~\cite{Tulyakov2018, zhu2020s3vae}.
For image generation, MUNIT~\cite{huang2018multimodal} disentangles the embedding of images into content feature and style feature. 
BicycleGAN~\cite{bicyclegan} combined cVAE-GAN~\cite{vaegan} and cLR-GAN~\cite{InfoGAN,bigan} to encourage the bijective consistency between the latent code and the output so that the model could generate different output by sampling different codes. 
For video generation, MoCoGAN~\cite{Tulyakov2018} and S3VAE~\cite{zhu2020s3vae} disentangle the motion from the object to generate videos in which different objects perform similar motions.
Different from~\cite{Tulyakov2018,zhu2020s3vae}, our method disentangle the motion representation into the audio-motion shared information and motion-specific information to model the one-to-many mapping between audio and motion.
\section{Preliminaries} \label{sec:motionvae}
In this section, we first briefly introduce the variational autoencoder (VAE)~\cite{kingma2014auto}, which is a widely used generative model. Then we describe 3D motion data and the most commonly used motion losses.
\subsection{Variational autoencoder}
Compared to autoencoder, VAE additionally imposes constraints on the latent code to enable sampling outputs from the latent space.
Specifically, during training, the distribution $P$ of the latent code is constrained to match a target distribution $Q$ with KL divergence as follows: 
\begin{equation}\small
\Scale[0.9]{\mathcal{D}\big(Q(z) \,\Vert\, P(z|X)\big) =  E_{z\sim Q}\big[\log{Q(z)}-\log{P(z|X)}\big]} ,
\end{equation}
where the $X$ represents the input of the corresponding encoder (audio or motion in our case), and $z$ represents its corresponding latent code.
The above goal can be achieved by minimizing the Evidence Lower Bound (ELBO) \cite{vaetutorial}:
\begin{equation}\small
\Scale[0.9]{\log{P(X|z)}-\mathcal{D}\big[Q(z|X)\,\Vert\, P(z)\big].} \label{eq:target}
\end{equation}

The second term of Eq.~\ref{eq:target} is a KL-divergence between two Gaussian distributions (with a diagonal covariance matrix). The prior distribution $P$ is set to Gaussian distribution (with a diagonal covariance matrix in our model, thus, the KL-divergence can be computed as:
\begin{equation}\small
\Scale[0.9]{\mathcal{D}=\frac{1}{2}\Big(\operatorname{tr}\big(\Sigma(X)\big)+\mu(X)^T\mu(X)-k-\log{\det\big(\Sigma(X)\big)}\Big)},
\end{equation}
where $k$ is the dimension of the distribution~\cite{vaetutorial}.
\subsection{Motion reconstruction loss}
In our method, the generated motion is supervised with {\it motion reconstruction loss}, consisting of rotation loss, position loss, and speed loss. 
Formally, it is defined as follows:
\begin{equation}\small
    \Scale[1]{L_{\text{mot}} = \lambda_{\text{rot}} \times L_{\text{rot}} + \lambda_{\text{pos}} \times L_{\text{pos}} + \lambda_{\text{speed}} \times L_{\text{speed}}}, \label{eq:mr}
\end{equation}
where $\lambda_{\text{rot}}$, $\lambda_{\text{pos}}$, $\lambda_{\text{speed}}$ are weights. We detail each term in the following.

Angular distance, i.e., geodesic distance, between the predicted rotation and the GT is adopted as the rotation loss.
Mathematically,
\begin{equation}
\Scale[0.9]{L_{\text{rot}} = \frac{1}{J\times T}\sum_{j=1}^J\sum_{t=1}^T cos^{-1}\frac{\operatorname{Tr}{\big(R_t^j(\hat{R}_t^j)^{-1}\big)} - 1}{2}}.
\end{equation}

Position loss is the $L_1$ distance between the predicted and target joint positions as follows:
\begin{equation}
\Scale[0.9]{    L_{\text{pos}}= \frac{1}{J\times T}\sum_{j=1}^J\sum_{t=1}^T\Vert\,\hat{p}_t^j-p_t^j\,\Vert_1}.
\end{equation}

Speed loss is introduced to help the model learn the complicated motion dynamics. 
In our work, the joint speed $v^j_t$ is defined as $v^j_t=p^j_{t+1} - p^j_t$.
We optimize the predicted and target joint speed as follows:
\begin{equation}
\Scale[0.9]{        L_{\text{speed}} = \frac{1}{J\times (T-1)}\sum_{j=1}^J\sum_{t=1}^{T-1}\Vert\,\hat{v}_t^j - v_t^j\,\Vert_1}.
\end{equation}

Our model can be trained with 2D motion data or 3D motion data.
When modeling the 2D human motion, our method directly predicts the joint position.
When modeling the 3D human motion, our method predicts the joint rotation and calculates the 3D joint positions with forward kinematics (FK).
Concretely, the FK equation takes in as input the joint rotation matrix about its parent joint and the relative translation to its parent joint (i.e. bone length) and outputs joint positions as follows:
\begin{equation}\small
p^j_t = p^{\text{parent}(j)}_t + R^j_ts^j ,
\label{eq:fk}
\end{equation}
where $R^j_t$ represents the rotation matrix of joint $j$ in frame $t$,
$p^j_t$ represents the position of joint $j$ in frame $t$,
$s^j$ represents the relative translation of joint $j$ to its parent,
and $\text{parent}(j)$ represents the parent joint index of the joint $j$.
We will always use $j$ and $t$ to index joints and frames in the following.
Our model predicts joint rotation in 6D representation~\cite{Zhou_2019_CVPR}, which is a continuous representation that help the optimization of the model.
The representation is then converted to rotation matrix $R^j_t$ by Gram-Schmidt-like process, where $R^j_t$ is the rotation matrix of joint $j$ in frame $t$.

\vspace{-0.1cm}
\section{Audio2Gestures}
\vspace{-0.1cm}
The proposed Audio2Gestures algorithm is detailed in this section. We first present our Audio2Gestures network by formulating the multimodal motion generation problem in Sec.~\ref{sec:net}, then we detail the training process in Sec.~\ref{sec:loss}.

\begin{figure*}[!t]
    \centering
    \begin{overpic}[trim=3.5cm 6.6cm 5cm 3.2cm,clip,width=1\linewidth,grid=false]{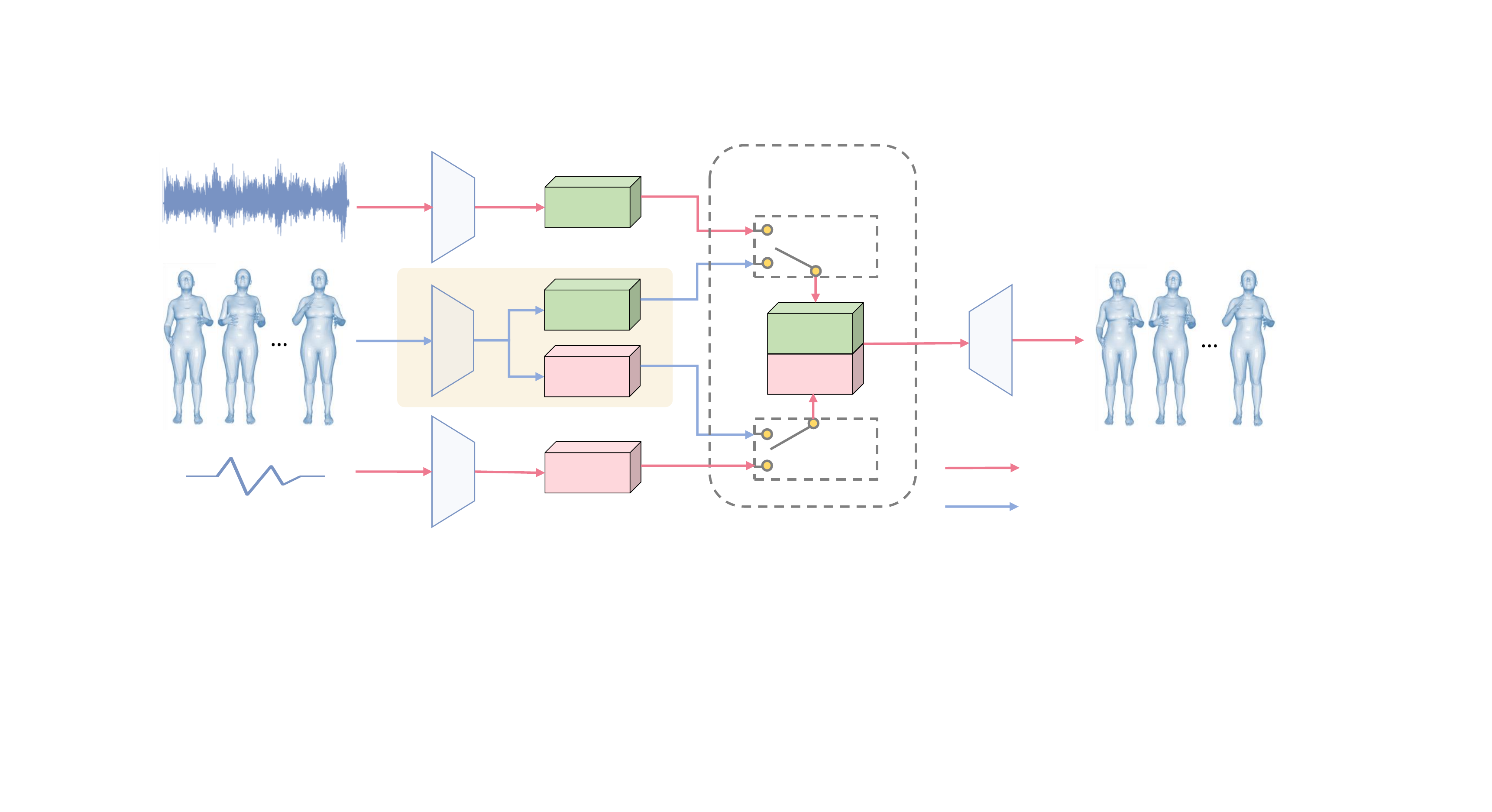}
    \put(2.5,1){\small Random Sampling}
    \put(25.5,4.5){\small $f_\text{R}$}
    \put(25.5,16.1){\small $f_\text{M}$}
    \put(25.5,28){\small $f_\text{A}$}
    \put(38,4.5){\small $I_\text{R}$}
    \put(38,13){\small $I_\text{M}$}
    \put(38,19){\small $S_\text{M}$}
    \put(38,28){\small $S_\text{A}$}
    \put(58,13){\small $I$}
    \put(58,16.6){\small $S$}
    \put(73.6,16.6){\small $g$}
    \put(51.5,30){\small Code Recombination}
    \put(78,5){\small Training/Inference data flow}
    \put(78,1.8){\small Training-only data flow}
    \end{overpic}
    \caption{Our method explicitly models the audio-motion mapping by splitting the latent code into shared and motion-specific codes. The decoder generates different motions by recombining the shared and motion-specific codes extracted from different sources. The data flow in blue is only used at the training stage because we do not have motion data during inference.}
    \label{fig:model}
\end{figure*}
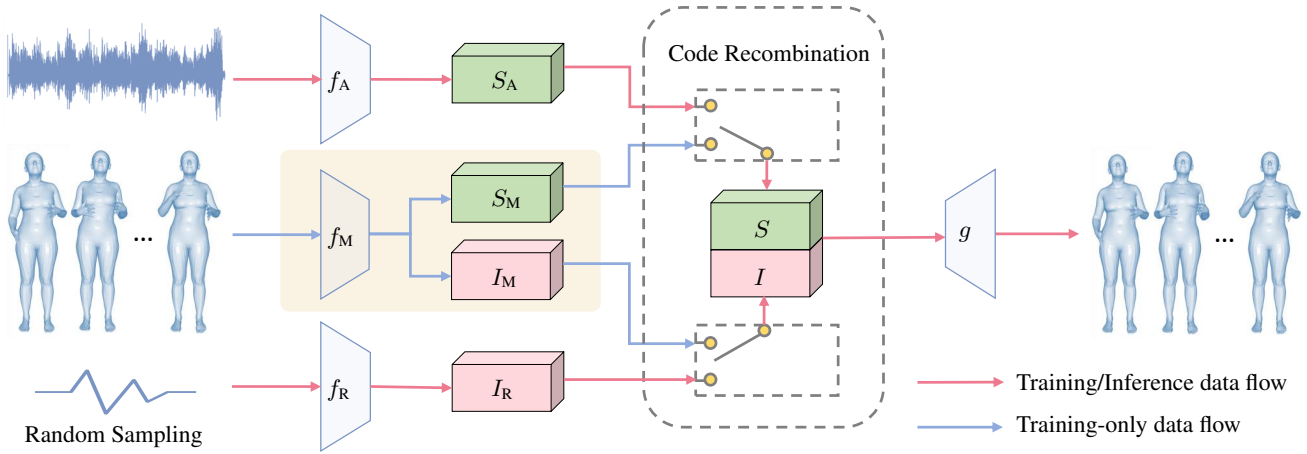

\subsection{Network structure} \label{sec:net}
We use a conditional encoder-decoder network to model the correlation between audio $A$ and motion $M = [p_1,p_2,...,p_T]$, where $p_t$ represents the joint positions of frame $t$.
In \reffig{fig:model}, our proposed model is made up of an audio encoder $f_A$, a motion encoder $f_M$, a mapping net $f_R$ to produce motion-specific code during inference, and a common decoder $g$ to generate motions from latent codes.
The latent code has been explicitly {\it split} into two parts (code $S$ and $I$) to account for the frequently occurred one-to-many mapping between the {\it same} (technically, very similar) audio and many different possible motions.
The mapping network is introduced to facilitate sampling motion-specific codes.
Under this formulation, given the same audio input (resulting in the same shared code $S_\text{A}$), varied motions produce different motion-specific code $I_\text{M}$ through motion encoder $f_\text{M}$, resulting in different {\it full} latent codes ($S_\text{A}\oplus I_\text{M}$) so that the network can better model the one-to-many mapping ($M = g(S_\text{A}, I_\text{M})$).
\par
During inference, shared feature $S_\text{A}$ is extracted with $f_\text{A}$ from the given audio $A$.
Motion-specific feature $I_\text{R}$ is generated with $f_\text{R}$ from a randomly sampled signal .
Both $S_\text{A}$ and $I_\text{R}$ are fed into the decoder $g$ to produce the final motion $M$, i.e $M = g(S_\text{A}, I_\text{R})$.
\par
During training, given a paired audio-motion data ${A}$ and ${M}$, their features ${S_\text{A}, S_\text{M}, I_\text{M}}$ are firstly extracted by the encoders. Concretely, $S_\text{A} = f_\text{A}(A)$ and $(S_\text{M}, I_\text{M}) = f_\text{M}(M)$.
The decoder learns to reconstruct the input motion from the extracted features.
To be specific, the decoder models the motion space by reconstructing the input motion by $\hat{M}=g(S_\text{M}, I_\text{M})$
The model is expected to learn the joint embedding of audio and motion by guiding the decoder generate the same target motion from shared codes extracted from different source.
But in practice, we notice the decoder will ignore the shared codes $S$ and reconstruct the motion only from $I_\text{M}$.
This is unwanted since the final motion is solely determined by the motion-specific features, being completely not correlated with the control signal (audio).
Thus, another data flow ($\hat{M}_{S_\text{A}I_\text{R}}=g(S_\text{A}, I_\text{R})$) is introduced so that the decoder has to utilize the information contained in the shared code extracted from audio to reconstruct the target.
The $I_\text{R}$ is generated from the mapping net $f_\text{R}$, whose input is a random signal from a Gaussian distribution.
The mean and variance of the distribution is calculated from the $I_\text{M}$ of the target motion per channel.
We experimentally find using a mapping network $f_\text{R}$ is helpful to improve the realism of the generated motions, which is mainly caused by the mapping network helps align the sampled feature with the motion-specific feature.
\vspace{-0.1cm}
\subsection{Latent code learning}  \label{sec:loss}
\vspace{-0.1cm}
\begin{figure}[!t]
\centering
\begin{overpic}[width=0.8\columnwidth,grid=false]{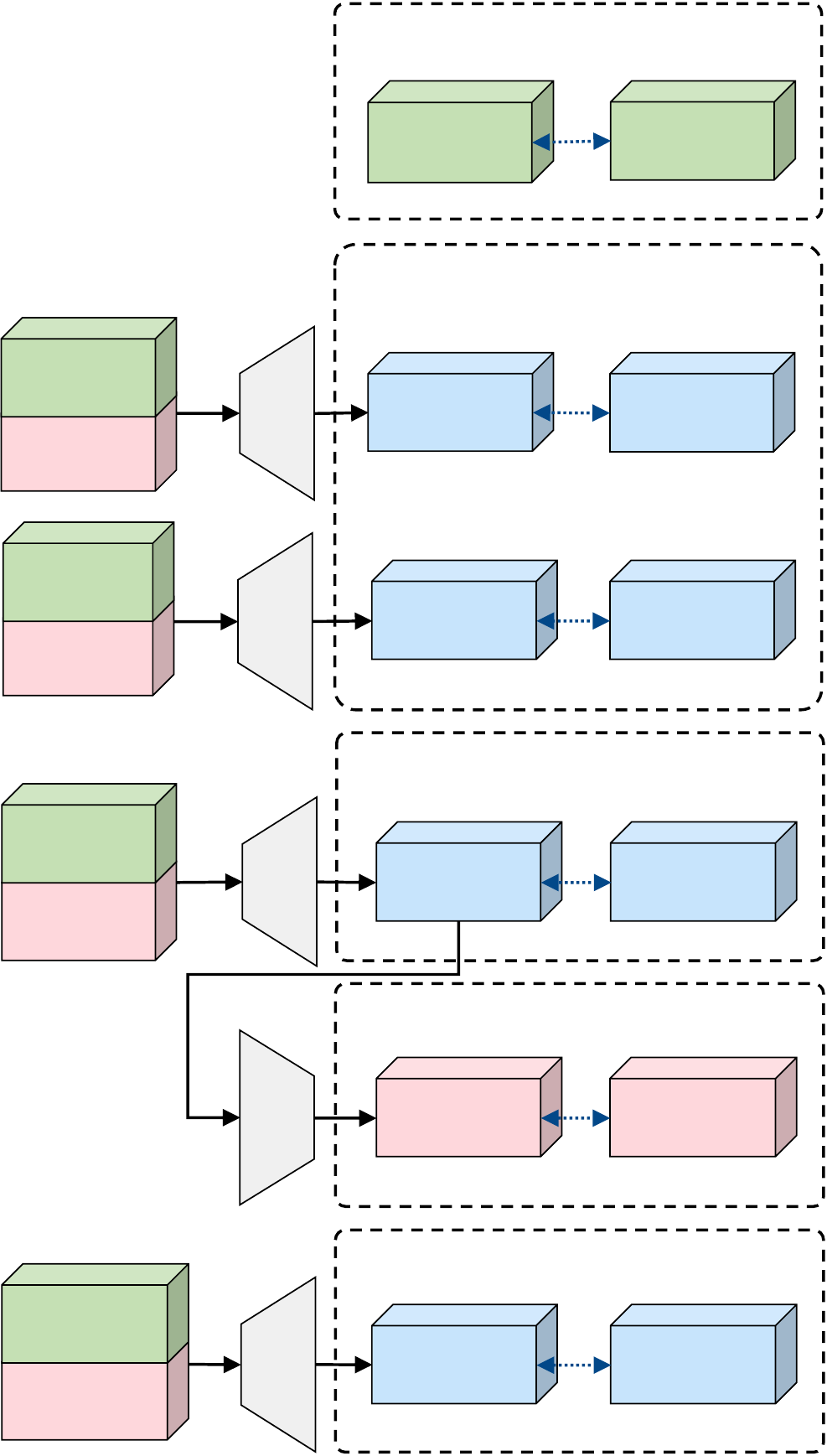}
\put(4,3){\small $I_{\text{R}_2}$}
\put(4,8){\small $S_{\text{A}}$}
\put(4,36){\small $I_{\text{R}_1}$}
\put(4,41){\small $S_{\text{A}}$}
\put(4,54){\small $I_{\text{M}}$}
\put(4,59){\small $S_{\text{A}}$}
\put(4,68){\small $I_{\text{M}}$}
\put(4,73){\small $S_{\text{M}}$}
\put(18.5,6){\small $g$}
\put(18.5,39){\small $g$}
\put(18.5,57){\small $g$}
\put(18.5,71){\small $g$}
\put(17,23){\small $f_{\text{M}}$}
\put(29,89){\small $S_{\text{M}}$}
\put(46,89){\small $S_{\text{A}}$}
\put(29,71){\small $\hat{M}$}
\put(46,71){\small $M$}
\put(27,57){\small $\hat{M}_{S_{\text{A}}I_{\text{M}}}$}
\put(46,57){\small $M$}
\put(27,39){\small $\hat{M}_{S_{\text{A}}I_{\text{R}_1}}$}
\put(46,39){\small $M$}
\put(29,22){\small $\hat{I}_{\text{R}_1}$}
\put(46,22){\small $I_{\text{R}_1}$}
\put(44,5.5){\small $\hat{M}_{S_{\text{A}}I_{\text{R}_1}}$}
\put(27,5.5){\small $\hat{M}_{S_{\text{A}}I_{\text{R}_2}}$}
\put(32,12){\small Diversity Loss}
\put(28,46){\small Relaxed Motion Loss}
\put(33,78){\small Motion Loss}
\put(30,29){\small Bicycle Constraint}
\put(28,96){\small Alignment Constraint}
\end{overpic}
\caption{The training details of our model. 
Our model is trained with alignment constraint, motion reconstruction losses, relaxed motion loss, bicycle constraint, diversity loss and KL divergence. 
The alignment constraints and motion reconstruction loss help the model learn the audio-motion joint embedding.
The relaxed motion loss avoids the degeneration of the shared code.
The bicycle constraints and the diversity loss help reduce the mode-collapse problem and guide the model to generate multimodal motions.
The KL divergence is omitted in the figure for the sake of brevity.
}
\label{fig:training}
\end{figure}
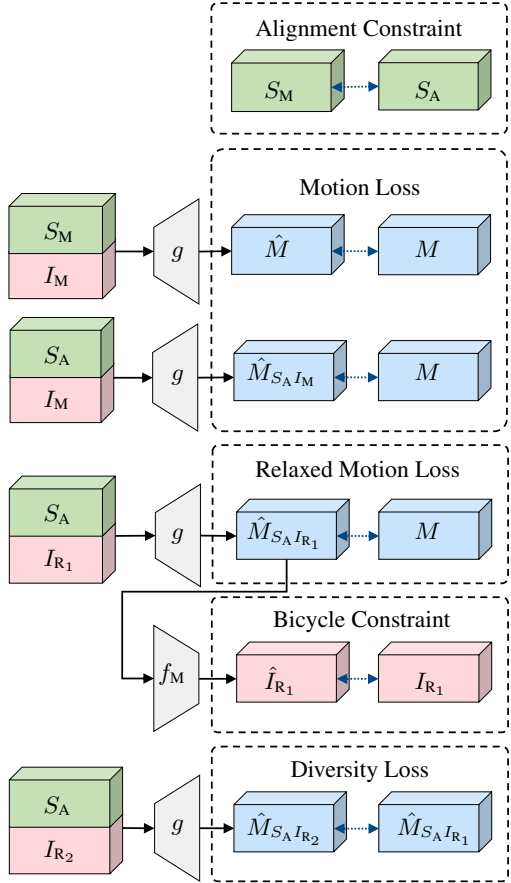

To better learn the split audio-motion shared and motion-specific latent codes, five types of losses are introduced (\reffig{fig:training}).
{\it Alignment constraint} and {\it relaxed motion loss} are introduced to learn the joint embedding (i.e., shared code) of the audio and motion. 
{\it Bicycle constraints} and {\it diversity loss} are introduced to model the multimodality of the motions.
{\it KL divergence} has been described in \refsec{sec:motionvae} and thus omitted.
The details are as follows. 

\vspace{0.1cm}
\noindent\textbf{Shared code alignment.}
The shared code of paired audio and motion is expected to be the same so that we can safely use audio-extracted shared code during inference and generate realistic and audio-related motions.
We align the shared code of audio and motion by the alignment constraint:
\begin{equation}\small
    L_\text{AC}= \Vert\, S_\text{A} - S_\text{M} \,\Vert_1. \label{eq:sca}
\end{equation}

\noindent\textbf{Degeneration avoidance.}
As we described in \refsec{sec:net}, the model easily results in the degenerated network, which means the shared code is completely ignored and has no effect on the generated motion. 
Our solution to alleviating such degeneration is introducing an extra motion reconstruction with audio extracted shared code $S_\text{A}$ and random motion-specific code $I_R$.
Ideally, the generated motion $\hat{M}_{S_\text{A}I_\text{R}}$ resembles its GT from some aspects but is not the same as its GT. In our case, We assume the generated poses are similar in the 3D world space.
Thus we propose {\it relaxed motion loss}, which calculates the position loss and penalizes the model only when the distance is larger than a certain threshold $\rho$:
\begin{equation}
\Scale[0.9]{    L_\text{S} = \frac{1}{J} \sum_{i=1}^J \max\big(\Vert\,\hat{p}_i - p_i \,\Vert_1-\rho, 0\big). \label{eq:rmr}}
\end{equation}

\noindent\textbf{Motion-specific code alignment.}
Although the motion-specific code could be sampled from Gaussian distribution directly, we noticed that the realism and diversity of the generated motions are not good.
The problem is caused by the misalignment of the Gaussian distribution and the motion-specific code distribution.
Thus, the mapping net is introduced to map the signal sampled from Gaussian space to the motion-specific embedding. 
At the training stage, we calculate the mean and variance for every channel and every sample of the $I_\text{M}$.
The sampled features will be fed into a mapping network, which is also a variational autoencoder, before concatenating them with different shared codes to generate motions.

\vspace{0.1cm}
\noindent\textbf{Motion-specific code reconstruction.} 
Although the model could model the multimodal distribution of audio-motion pair by splitting the motion code into audio-motion shared one and motion-specific ones, it is not guaranteed the decoder can sample multimodal motions. For example, suppose the mapping net only maps the sampled signal to a single mode of the multimodal distribution. In that case, the decoder still could only generate unimodal motions, which is also known as the mode-collapse problem.
The bicycle constraint~\cite{bicyclegan} ($M \rightarrow I \rightarrow M$ and $I \rightarrow M \rightarrow I$) is introduced to avoid the mode-collapse problem, which encourages a bijection between the motion and the motion-specific code.
Since the motion reconstruction loss has already been introduced, an extra reconstruction loss of the motion-specific code is added as supplement:
\begin{equation}\label{eq:cr}
\Scale[0.9]{    L_\text{cyc} = \Vert\, \hat{I}_\text{R} - I_\text{R} \,\Vert_1 }.
\end{equation}

\noindent\textbf{Motion diversification.}
To further encourage multimodality of the generated motion, diversity loss~\cite{MSGAN, Choi_2020_CVPR} is introduced.
Maximizing the multimodality loss encourages the mapping network to explore the meaningful motion-specific code space. We follow the setting in~\cite{Choi_2020_CVPR} and directly maximize the joint position distance between two sampled motions 
since it is more stable than the original one~\cite{MSGAN}:
\begin{equation}\label{eq:ds}
\Scale[0.9]{L_\text{DS}=-L_{\text{pos}}(\hat{M}_{S_{\text{M}}I_{\text{R}_1}}, M).}
\end{equation}

\section{Experiments}
In this section, we first introduce the datasets, evaluation metrics and implementation details separately in \refsec{sec:datasets}-\ref{sec:implementation}. 
Then we show the performance of our algorithm and compare it with three state-of-the-art methods \ref{sec:sota}. 
Finally, we analyze the influence of each module of our model on the performance by ablation studies \ref{sec:abla}.
More results are presented in our project page\footnote{\url{https://jingli513.github.io/audio2gestures}}.

\subsection{Datasets} \label{sec:datasets}
\noindent\textbf{Trinity dataset.}
Trinity Gesture Dataset~\cite{IVA:2018} is a large-scale speech to gesture synthesis dataset. This dataset records a male native English speaker talking many different topics, such as movies and daily activities. 
The dataset contains 23 sequences of paired audio-motion data, 244 minutes in total. 
The audio of the dataset is recorded at 44kHz.
The motion data, consisting of 56 joints, are recorded at 60 frame per second (FPS) or 120 FPS using Vicon motion capture system.

\vspace{0.1cm}
\noindent\textbf{S2G-Ellen dataset.}
The S2G-Ellen dataset, which is a subset of the Speech2Gesture dataset~\cite{Ginosar2019}, contains positions of 49 2D upper body joint estimated from 504 YouTube videos, including 406 training sequences (469513 frames), 46 validation sequences (46027 frames), and 52 test sequences (59922 frames).
The joints, which is estimated using OpenPose~\cite{cao2017realtime}, include neck, shoulders, elbows, wrists, and hands.

\subsection{Evaluation metrics}\label{sec:eval}
\subsubsection{Quantitative metrics}
\noindent\textbf{Realism.} Following Ginosar \textit{et al.}'s~\cite{Ginosar2019} suggestion, the $L_1$ distance of joint position in Eq.~\ref{eq:l1} and the percentage of correct 3D keypoints (PCK) in Eq.~\ref{eq:PCK} are adopted to evaluate the realism of the generated motion.
Specifically, $L_1$ distance is calculated by averaging the corresponding joint's position error of all joints between prediction $\hat{M}$ and GT $M$:
\begin{equation}\label{eq:l1}
\Scale[0.9]{    L_1 = \frac{1}{T\times J}\sum_{t=1}^T \sum_{j=1}^J\Vert\hat{M}-M\Vert_1 }.
\end{equation}
The PCK metric calculates the percentage of correctly predicted keypoints, where a predicted keypoint is thought as correct if its distance to its target is smaller than a threshold $\delta$:
\begin{equation}\label{eq:PCK}
    \Scale[0.9]{   \text{PCK} = \frac{1}{T\times J}\sum_{t=1}^T \sum_{j=1}^J{\mathbf{1}\big[\Vert p^j_t - \hat{p}^j_t\Vert_2<\delta\big]}},
\end{equation}
where $\mathbf{1}$ is the indicator function and $p^j_t$ indicates joint $j$'s position of frame $t$.
As in~\cite{Ginosar2019}, the $\delta$ is set to 0.2 in our experiments.
\par
\vspace{0.1cm}
\noindent\textbf{Diversity.} 
Diversity measures how many different poses/motions have been generated within a long motion.
For example, RNN-based methods easily get stuck to some static motion as the generated motion becomes longer and longer. And static motions, which are undesired apparently, should get low diversity scores.
We first split the generated motions into equal-lengthed non-overlapping motion clips (50 frames per clip in our experiments) and we calculate diversity as the averaged $L_1$ distance of the motion clips.
Formally, it is defined as:
\begin{equation}\label{eq:diversity}
    \Scale[0.9]{\text{Diversity}=\frac{1}{N \times \ceil{N/2}} \sum_{a_1=1}^{N} \sum_{a_2=a_1+1}^{N}\Vert\hat{M}_{a_1} - \hat{M}_{a_2}\Vert_1},
\end{equation}
where the $\hat{M}_{a_1}$ and $\hat{M}_{a_2}$ represent clips from the same motion sequence, $N$ represents the count of the motion clips, which is $\frac{T}{50}$ in our experiments.
Please note that jitter motion and invalid poses can also result in high diversity score. So higher diversity is preferred only if the generated motion is natural.
\par
\vspace{0.1cm}
\noindent\textbf{Multimodality.}
Multimodality measures how many different motions could be sampled (through multiple runs) for a given audio clip.
Note that multimodality calculates motion difference across different motions while diversity calculates (short) motion clip difference within the same (long) motion.
We measure the multimodality by generating motions for an audio $N$ times, which is 20 in our experiments, and then calculate the average $L_1$ distance of the motions.
\begin{equation}
    \Scale[0.9]{\text{Multimodality} = \frac{1}{N \times \ceil{N/2}} \sum_{a=1}^N \sum_{b=a+1}^N\Vert\hat{M}_a-\hat{M}_b\Vert_1},
\end{equation}
where the $\hat{M}_a$ and $\hat{M}_b$ represent
sampled motions generated through different runs for the given audio.
Similar to diversity, invalid motion will also result in abnormally high multimodality score.
\vspace{-0.1cm}
\subsubsection{User studies}
\begin{figure}[!t]
\centering
\includegraphics[width=\columnwidth,trim=0.3cm 15cm 0cm 0cm,clip]{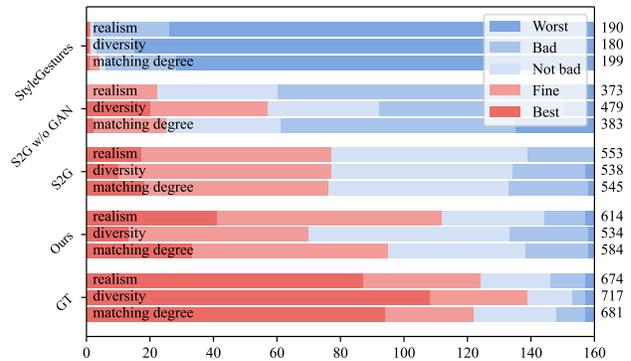}
\DeclareGraphicsExtensions.
\vspace{-0.3cm}
\caption{
    User study results comparing our method against the state-of-the-art methods. ``S2G'' is short for Speech2Gesture~\cite{Ginosar2019}. 
    The horizontal axis represents the number of samples rated by the participants. In total, 160 comparisons have been rated (40 participants, 4 comparisons each questionnaire).
    The average score (higher is better) for each method is listed on the right.
    Bars with different colors indicate the count of the corresponding ranking of each algorithm.
    The video results are in our project page. 
    }
\label{fig:user_study}
\end{figure}

\vspace{-0.1cm}
To evaluate the results qualitatively, we conduct user studies to analyze the visual quality of the generated motions.
Our questionnaire contains four 20-second long videos. The motion clips shown in one video is generated by various methods from the same audio clip.
The participants are asked to rate the motion clips from the following three aspects respectively: 
\begin{enumerate}[topsep=0pt,itemsep=0pt,parsep=0pt,partopsep=0pt]
\item Realism: which one is more realistic?
\item Diversity: which motion has more details?
\item Matching degree: which motion matches the audio better?
\end{enumerate}
The results of the questionnaires are shown in \reffig{fig:user_study}.
We show the count of different ranking in the figure. The average score of different metrics for each algorithm is listed after the corresponding bar.
The scores assigned to each ratings are \{5,4,3,2,1\} for \{best, fine, not bad, bad, worst\} respectively.

\begin{table*}[!t]
\centering
\small
{\def\arraystretch{0.8} \tabcolsep=1.8em 
\begin{tabular}{ll|r@{\hspace{0.1cm}}lr@{\hspace{0.1cm}}lr@{\hspace{0.1cm}}lc}
\toprule[1pt]
Dataset & Method & \multicolumn{2}{c}{$L_1\downarrow$}  & \multicolumn{2}{c}{PCK $\uparrow$} & \multicolumn{2}{c}{Diverisity $\uparrow$ }& Multimodality $\uparrow$ \\
\midrule[1pt]
\multirow{4}{*}{Trinity} & S2G w/o GAN \cite{Ginosar2019} & 7.71 &  & 0.82 & & 5.99&  & - \\
& S2G \cite{Ginosar2019} & 24.68 &  & 0.39 &  & 2.46 & &  - \\
& StyleGestures \cite{stylegestures} & 18.97 & (18.07) & 0.34 & (0.34) & 2.34 & (3.79) & \textbf{7.55} \\
& Ours & \textbf{7.84} & \textbf{(7.65)} & \textbf{0.82} & \textbf{(0.83)} & \textbf{6.32} & \textbf{(6.52)} & 4.11 \\
\midrule[0.8pt]
\multirow{3}{*}{S2G-Ellen} & S2G w/o GAN \cite{Ginosar2019} & 0.74 & & 0.37 &  & 0.61 &  & - \\
& S2G \cite{Ginosar2019} & 1.08 &  & 0.23 &  & 0.89 & & - \\
& Ours & 0.94 & (0.92) & 0.33 & (0.34) & 0.84 & (0.85) & 0.77 \\
\bottomrule[1pt]
\end{tabular}}
\caption{Quantitative results on Trinity dataset and S2G-Ellen dataset. $\uparrow$ means the higher is better and $\downarrow$ means the lower is better. For methods supporting sampling, we run 20 tests and report their {\it average} score and the {\it best} score (in parentheses). Speech2Gesture (``S2G'' in the table) could not generate multimodality motions.}
\label{tab:sota_result}
\end{table*}

\begin{table*}[!t]
\small
\centering
{\def\arraystretch{0.8}\tabcolsep=2.7em 
  \begin{tabular}{l|r@{\hspace{0.1cm}}lr@{\hspace{0.1cm}}lr@{\hspace{0.1cm}}lc}
       \toprule[1pt]
      Method & \multicolumn{2}{c}{$L_1$ $\downarrow$}  & \multicolumn{2}{c}{ PCK $\uparrow$ } &\multicolumn{2}{c}{ Diversity $\uparrow$} & Multimodality $\uparrow$ \\
       \midrule[1pt]
      baseline & 8.22 & & 0.80 & & 6.20 & & - \\
      \hspace{0.1cm}+ split & 8.69 & (8.30) &0.77 & (0.78) & 5.83 & (6.02) & \textbf{5.90} \\
      \hspace{0.1cm}+ mapping net & 8.06 & (7.91) & 0.80 & (0.81) & 5.86 & (6.05) & 3.44 \\
      \hspace{0.1cm}+ bicycle constraint & 7.94 & \textbf{(7.63)} & 0.80 & (0.82) & 6.31 & (6.46) & 3.68\\
      \hspace{0.1cm}+ diversity loss & \textbf{7.84} & (7.65) & \textbf{0.82} & \textbf{(0.83)} & \textbf{6.32} & \textbf{(6.52)} & 4.11\\
      \bottomrule[1pt]
   \end{tabular}}
\caption{Ablation study results on the Trinity dataset. Note that every line adds a new component compared to its previous line. For methods supporting sampling, we run 20 tests and report their {\it average} score and the {\it best} score (in parentheses).}
\label{tab:3D_ablation_study}
\end{table*}
\subsection{Implementation details}
\label{sec:implementation}
\noindent\textbf{Data processing.} \label{sec:data_proc}
We detail the data processing of Trinity dataset and S2G-Ellen dataset here.

(1) Trinity dataset.
The audio data are resampled to 16kHz for extracting log-mel spectrogram~\cite{stevens1937scale} feature using librosa~\cite{brian_mcfee_2020_3955228}.
More concretely, the hop size is set to $ \text{SR} / \text{FR} $ where $\text{SR}$ is the sample rate of the audio and FR is the frame rate of the motion so that the resulting audio feature have the same length as the input motion.
In our case, the resulting hop size is 533 since $\text{SR}$ is 16000 and $\text{FR}$ is 30.
The dimension of the log-mel spectrogram is 64.
\par
The motion data are downsampled to 30 FPS and then retargeted to the SMPL-X~\cite{SMPL-X:2019} model.
SMPL-X is an expressive articulated human model consisting of 54 joints (21 body joints, 30 hand joints, 3 face joints, respectively) , which has been widely used in 3D pose estimation and prediction~\cite{hmr,SMPL-X:2019,phd,spin}.
The joint rotation is in 6D rotation representation~\cite{Zhou_2019_CVPR} in our experiments, which is a smooth representation and could help the model approximate the target easier. 
Note that the finger motions are removed due to unignorable noise.

(2) S2G-Ellen dataset. Following~\cite{Ginosar2019}, the data are split into 64-frame long clips (4.2 seconds). 
Audio features are extracted in the same way as the Trinity dataset.
The body joints are represented in a local coordinate frame relative to its root. Namely, the origin of the coordinate is the root joint.

\vspace{0.1cm}
\noindent\textbf{Network.}
Every encoder, decoder and mapping net consists of four residual blocks~\cite{he2016deep}, including 1D convolution and ReLU non-linearity~\cite{agarap2018deep}. 
The residual block is similar to~\cite{TCN} except several modifications. 
To be specific, the casual convolutions whose kernels see only the history are replaced with normal symmetric 1D convolutions seeing both the history and the future.
Both the shared code and motion-specific code are set to 16 dimensions. 

\vspace{0.1cm}
\noindent\textbf{Training.}
At the training stage, we randomly crop a 4.2-second segment of the audio and motion data, which is 64 frames for the S2G dataset (15 FPS) and 128 frames for the Trinity dataset (30 FPS).
The model weights are initialized with the Xavier method~\cite{glorot2010understanding} and trained 180K steps using the Adam~\cite{kingma2014adam} optimizer. 
The batch size is 32 and the learning rate is $10^{-4}$.
The $\lambda_{\text{rot}}$, $\lambda_{\text{pos}}$, $\lambda_{\text{speed}}$ are set as ${1,1,5}$ respectively,
and $\rho$ is set as $0.02$ in our experiments.
Our model is implemented with PyTorch~\cite{paszke2017automatic}.
\subsection{Comparison with state-of-the-art methods} \label{sec:sota}
We compare our method with two recent representative state-of-the-art methods, including one LSTM-based method named StyleGestures~\cite{stylegestures} and one CNN-based method named Speech2Gesture~\cite{Ginosar2019} on Trinity dataset.
StyleGestures adapts normalizing flows~\cite{kobyzev2020normalizing,kingma2018glow,henter2019moglow} to speech-driven gesture synthesis.
We train StyleGestures using the code released by the authors. 
The training data of the StyleGestures are processed in the same way as the authors indicate\footnote{The motions generated by StyleGestures are 20 FPS and have a different skeleton from our method. We upsample the predicted motion to 30 FPS and retarget it to SMPL-X skeleton with MotionBuilder.}.
Speech2Gesture, originally designed to map speech to 2D human keypoints, consists of an audio encoder and a motion decoder. 
Its final output layer has been adjusted to predict 3D joint rotations and is trained with the same losses as our method. 

Quantitative experimental results are listed in \reftab{tab:sota_result} and user study results in \reffig{fig:user_study}.
Both results show that our method outperforms previous state-of-the-art algorithms on the realism and diversity metrics, demonstrating that it is beneficial to explicitly model the one-to-many mapping between audio and motion in the network structure.

While StyleGestures supports generating different motions for the same audio by sampling, the quality of its generated motions is not very appealing.
Also, its diversity score is the lowest, because LSTM output easily gets stuck into some poses, resulting in long static motion afterwards.
The algorithm is not good at generating long-term sequences due to the error accumulation problem of the LSTM.
The authors test their algorithm on 400 frames (13 seconds) length sequences.
However, obviously deteriorated motions are generated when evaluating their algorithm to generate 5000-frame (166 seconds) long motions.

As for Speech2Gesture, the generated motions show similar realism with ours but obtain lower diversity score (\reftab{tab:sota_result}) than our method.
But Speech2Gesture does not support generating multimodal motions.
Also note that Speech2Gesture with GAN generates many invalid poses and gets the worst performance.
We have trained the model several times changing the learning rate range from 0.0001 to 0.01, and report the best performance here.
The bad performance may be caused by the unstable of the training process of the generative adversarial network.
\subsection{Ablation study} \label{sec:abla}
To gain more insights into the proposed components of our model, we test some variants of our model on the 3D Trinity dataset (\reftab{tab:3D_ablation_study}).
We run every variant 20 times and report the {\it averaged} performance and the {\it best} performance to avoid the influence of randomness.
Note that the randomness of our model comes from two different parts, the randomness introduced by the variational autoencoder and by the motion-specific feature sampling.
\par
We start with a ``baseline'' model, which excludes the mapping net and the split code. It is trained only with the motion reconstruction losses (Eq.~\ref{eq:mr}) and shared code constraint (Eq.~\ref{eq:sca}).
The averaged scores ``avg $L_1$'', ``avg PCK'' and ``avg Diversity'' of the model equal to the best scores ``min $L_1$'', ``max PCK'' and ``max Diversity'' on $L_1$, which indicates that the randomness of the VAE model have almost no affect on generating multimodal motions.
\par
The next setting is termed as ``+split'', which splits the output of the motion encoder into shared and motion-specific codes and introduces the relaxed motion loss (Eq. \ref{eq:rmr}). 
This modification explicitly enables the network to handle the one-to-many mapping, but it harms the realism (see ``avg $L_1$'', ``min $L_1$'', ``avg PCK'' and ``max PCK'') and diversity. 
As we can see, both the $L_1$ and the PCK metrics are worse than ``baseline''.
The abnormal results is mainly caused by the misalignment between the sampled signal and the motion-specific feature.
We analyze the difference between the sampled signals with the motion-specific feature, and find that there is a big difference in their statistical characteristics, such as the mean and variance of derivatives.
\par
Thus, a mapping network (``+mapping net'') is introduced to align the sampled signal with the motion-specific feature automatically.
Although the multimodality drops compare to ``+split'', this modification helps to improve other metrics of the generated motions a lot.
Note that a higher multimodality score only makes sense when the generated motions is natural, as described in \refsec{sec:eval}.
The ``+mapping net'' model also outperforms the baseline model in the $L_1$ metrics and gets a similar PCK metrics, but the model could generate multimodal motions.
We notice that the diversity of the motions generated by ``+mapping net'' model is not as good as the baseline model.
The realism score of the ``+mapping net'' model is also worse than the baseline model, which may be due to the users prefer the motions with more dynamics.
We think the problem may be caused by the mode collapse problem suffered by many generative methods.

To overcome this problem, two simple yet effective losses -- Bicycle constraints and diversity loss -- are introduced.
Bicycle constraint improves the multimodality of the motions from 3.44 to 3.68. The avg diversity of the motions also increase from 5.86 to 6.31.
The diversity loss further improves the motion diversity and multimodality but have little influence on the realism.
The final model outperforms the baseline model in all quantitative indicators, which shows that the audio-motion mapping could be better modeled by explicit modeling the one-to-many correlation.
\subsection{Application}
We notice that motion-specific code extracted from a motion strongly controls the final motion output.
To be specific, the synthesized motion is almost the same as the original motion used to extract this motion-specific code.
This feature is perfect for a type of motion synthesis application where pre-defined motions are provided on the timeline as constraints.
For example, if there is a $n$-frame long motion clip that we want the avatar to perform from frame $t$ to $t+n$.
We could extract its motion-specific code $I_M$ with the motion encoder and directly replace the sampled motion-specific code $I_R$ from $t$ to $t+n$.
Our model could generate a smooth motion from the edited motion-specific code. 
Please refer to our project page for the demonstration.

\section{Conclusion}
In this paper, we explicitly model the one-to-many mapping by splitting the latent code into shared code and motion-specific code. 
This simple solution with our customized training strategy effectively improves the realism, diversity, and multimodality of the generated motion.
We also demonstrate an application that the model could insert a specific motion into the generated motion by editing the motion-specific code, with smooth and realistic transitions.
Despite the model could generate multimodal motions and provide users the ability to control the output motion, there exist some limitations. For example, the generated motion is not very related to what the person says, future work could be improving the meaning of the generated motion by incorporating word embedding as an additional condition.
\section{Acknowledgement}
This work was supported by the Natural Science Foundation of China (U2013210, 62006060), the Shenzhen Research Council (JCYJ20210324120202006), the Shenzhen Stable Support Plan Fund  for Universities (GXWD20201230155427003-20200824125730001), and the Special Research project on COVID-19 Prevention and Control of Guangdong
Province (2020KZDZDX1227).
\clearpage

{\small
\normalem
\bibliographystyle{ieee_fullname}
\bibliography{audio2gestures_iccv_final}
}

\end{document}